\documentclass{ecai}
\usepackage{times}
\usepackage{graphicx}
\usepackage{latexsym}

\usepackage{subfigure}
\usepackage[section]{placeins}
\usepackage{amsmath}
\ecaisubmission   

\begin{document}

\title{Angular Learning: \\
Toward Discriminative Embedded Features}
\author{JT Wu \institute{University of Jinan, Jinan 250022, China, email: clouderow@gmail.com} \and L. Wang}

\maketitle
\bibliographystyle{ecai}

\begin{abstract}
The margin-based softmax loss functions greatly enhance intra-class compactness and perform well on the tasks of face recognition and object classification. Outperformance, however, depends on the careful hyperparameter selection. Moreover, the hard angle restriction also increases the risk of overfitting. In this paper, angular loss suggested by maximizing the angular gradient to promote intra-class compactness avoids overfitting. Besides, our method has only one adjustable constant for intra-class compactness control. We define three metrics to measure inter-class separability and intra-class compactness. In experiments, we test our method, as well as other methods, on many well-known datasets. Experimental results reveal that our method has the superiority of accuracy improvement, discriminative information, and time-consumption.
\end{abstract}

\section{INTRODUCTION}
\begin{figure}[!t]
  \centering
  \includegraphics[width=.45\textwidth]{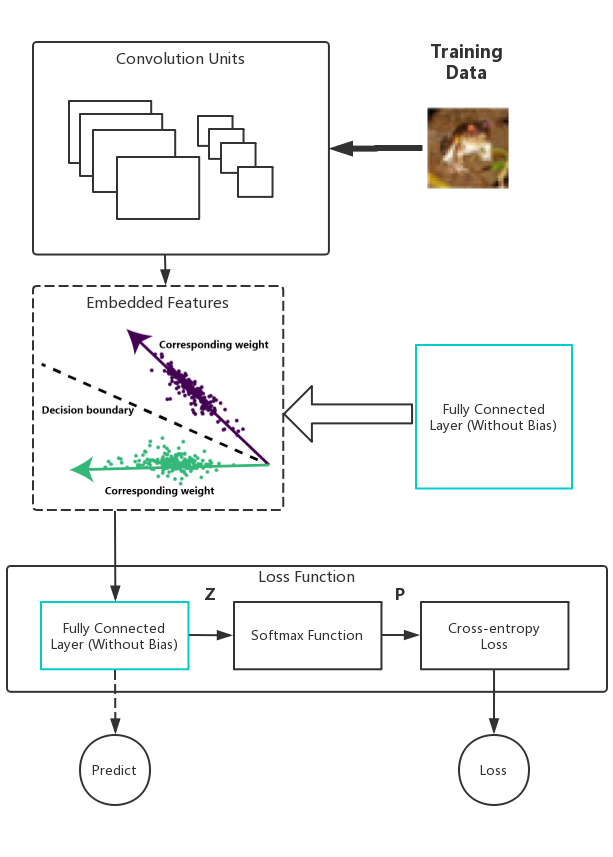}
  \caption{The framework of convolution representation learning. In representation learning, the weights of the last fully connected layer, also seemed as a part of loss function, help the convolution units learn good feature embedding.}
  \label{fig:framework}
\end{figure}

Traditional representation learning methods, like  Locally Linear Embedding (LLE) \cite{roweis2000nonlinear}, Laplacian Eigenmap \cite{belkin2002laplacian} and  Hessian LLE \cite{donoho2003hessian}, can not produce a meaningful feature embedding or distance metric. More important, these traditional methods can not deal with unseen samples. A neural network with a single layer is linear separability. Hence, the outputs of the second last fully connected layer are excellent features for distinguishing classes. representation learning by Deep Convolutional Neural Network (DCNN) embedding can generate embedded features for unseen samples and achieves excellent performance on face recognition tasks \cite{sun2014deep,sankaranarayanan2016triplet,taigman2014deepface}.  Figure \ref{fig:framework} illustrates the general framework of representation learning by DCNN. The softmax function following cross-entropy loss is the most common component in DCNN on classification tasks. However, the learned features are separable but not discriminative enough. This shortage raises the risk of learned features being misclassified near the decision boundary. Many researchers have noticed that by designing specific loss functions, discriminative features will improve performance.

Contrastive loss \cite{hadsell2006dimensionality} and triplet loss \cite{schroff2015facenet} use data pairs to enhance intra-class compactness and enlarge inter-class separability. They emphasize the relationship between features but increase the number of training pairs, which are theoretically up to $O(N^2)$ for the contrastive loss and to $O(N^3)$ for the triplet loss (where $N$ is the size of the training set). Additionally, the complex form of these loss functions makes it hard to train.
To encourage the intra-class compactness with low computational overhead, center loss \cite{wen2016discriminative} converges the learned features to their corresponding class centers. However, experiments (see in Table [\ref{table:as1},\ref{table:as2}]) show that this method declines the inter-class separability.

A study line of margin-based methods \cite{wang2018additive,wang2018cosface,wang2018soft-margin}, has several good properties, including clear geometric interpretation, strong intra-class compactness and easy to implement. {They also demonstrate the power of margin in representation learning. Margin is a hard constraint which enforces features belonging to different classes far away.} ArcFace, the outstanding one on margin-based methods, defines an addictive angular margin parameter $m$ to enforce the intra-class angle smaller than the inter-class angle significantly, $\theta_{y_i}<\mathrm{max}(\theta_j)-m$. Nevertheless, the high performance needs a careful selection of hyperparameters, and {the tiny intra-class angles may cause overfitting.}

In this paper, an {angular loss} is proposed to generate discriminative features. Our objective is that a loss function with a high angular gradient leads to a low intra-class angle. We also argue that margin-based methods face the risk of overfitting---for some cases, the feature of a mislabeled or high variance sample, forced into the center of this class, declines generalization ability. Our method has not this kind of hard constraint on features, allowing outliers separable. Without losing flexibility, our method also provides a hyperparameter $s$ to adjust the intra-class compactness. The summary of our contributions in this work is the following:
\begin{itemize}
    \item We cast a new viewpoint on the weakness of the softmax loss. i.e., the angular gradient of the softmax loss is closing to zero while the intra-class features are converging to the corresponding center.
    \item We propose a novel loss function, angular loss, which has an intra-class compactness regulation that is elastic. The approach promotes the angle small by the angular gradient in a soft way, rather than providing a restriction on the angles in a hard way like margin-base methods. 
    \item Experimental results on datasets of Fashion-MNIST, CIFAR10, and CIFAR100 reveal the effectiveness of our method.
\end{itemize}

In Section 2, we describe the preliminary knowledge and the methods we are going to compare. Section 3 introduces our method and the angular gradient. Empirically, our method consistently outperforms other methods on the accuracy, intra-class compactness, and computational efficiency, as shown in Section 4. For clarity, our work demonstrates a novel, convenient, and adjustable way to achieve intra-class compactness.

\section{RELATED WORKS}
\subsection{Softmax loss}
In supervised representation learning by DCNN, the softmax loss obtains a fully connected layer, a softmax function, and a cross-entropy loss function, defined by \cite{liu2016large}, shown in Figure \ref{fig:framework}. The representation of data learned by the deep convolution units usually transforms high dimension feature to low dimension embedded feature. And the class probabilities of a sample $x$, symbolized by $P(c)$, is calculated by the softmax function:
$$P(c=i|f;x)={\displaystyle {\frac {e^{f(x)_{i}}}{\sum _{j=1}^{C}e^{f(x)_{j}}}}{\text{ for }}i=1,\dotsc ,C}$$ where $C$ is the number of classes and $f$ represents the DCNN.
Then we get the loss by the cross-entropy loss,
$$L=-\log(P(c=y)),$$
where $y$ is the ground truth label for the sample $x$. 
To better describe the following methods, in this paper, we define that $Z$ is the output of the fully connected layer, $P$ is the output of the softmax function, and L is the final output of the loss function. For convenience, we denote the output of the convolution units or embedded feature just by $x$. And the superscript of those symbols distinguishes between different loss functions.\\

In the softmax loss, the last fully connected layer calculates the similarity between the feature $x$ and class $c$ by dot product, $Z^1_c=W^T_c\ x + b_c$, where $W$ are the weight matrix of this layer.  Finally, the original softmax loss can be written  as:
\begin{eqnarray}
  L^{\mathrm{original\_soft}}  =-\frac{1}{N} \sum_{i=1}^N \log\frac{e^{W^T_{y_i} x_i + b_{y_i}}}{\sum_{j=1}^Ce^{W^T_{j} x_i+b_j}}.
\end{eqnarray}
where $x_i$ is the embedded feature of $i$-th sample, $N$ is the number of a mini-batch size and $y_i$ is the lable of $i$-th sample. SphereFace \cite{liu2017sphereface} studies the effects of bias $b$. They find that omitting the bias does no harm to the performance but makes it easy to analyze. So we follow this modified softmax loss:
\begin{eqnarray}
  \label{eq:soft}
  \begin{split}
    L^{\mathrm{soft}} &= -\frac{1}{N} \sum_{i=1}^N \mathrm{log}\frac{e^{W^T_{y_i} x_i}}{\sum_{j=1}^Ce^{W^T_{j} x_i}}\\
    &=-\frac{1}{N} \sum_{i=1}^N \mathrm{log}\frac{e^{\left \| W^T_{y_i} \right \| \left \| x_i \right \| \mathrm{cos}(\theta_{y_i})}}    {\sum_{j=1}^Ce^{\left \| W^T_{j} \right \| \left \| x_i \right \| \cos(\theta_j)}}
  \end{split}
\end{eqnarray}
where $\theta$ is the angle between two vectors; $\theta_j$ denotes the angle between this feature and the weight vector of class $j$ or $W_j$.

The traditional softmax loss is the standard method on classification task. However, recent studies show that this method cannot generate discriminative features. The features near the decision boundary are more likely to be misclassified. To be discriminative features helps the ability of generalization on unseen samples. 

\subsection{ArcFace}
Weight normalization \cite{wen2016discriminative} can not only accelerate training but also solve the imbalanced data problem by rebalancing the weight of each class \cite{y2017oneshot}. 
Following \cite{liu2017sphereface,wang2018cosface,Wang_2017NormFace}, we fix the weight $\left \| W_j \right \|=1, \left \| x_i \right \|=1$ by $l_2$ normalisation and multiplied a rescale hyperparameter $s$, so the formulation  becomes
\begin{eqnarray}
  \label{eq:soft_norm}
  \begin{split}
    L^{\mathrm{cosine}}=-\frac{1}{N} \sum_{i=1}^N log\frac{e^{s\ \cos(\theta_{y_i})}}
    {e^{s\ \cos(\theta_{y_i})}  +\sum_{j=1,j\neq y_i}^C s\ e^{\cos(\theta_j)}}.
  \end{split}
\end{eqnarray}

L-softmax \cite{liu2016large} provides a new way, incorporating margin, to enhance intra-class compactness and inter-class separability simultaneously. ArcFace, L-softmax and soft-margin \cite{wang2018soft-margin} use different margin penalty and achieve good results, however, ArcFace was proved outstanding on them \cite{deng2019arcface}. ArcFace applies an angular margin $m$ to penalise the angle:$cos(\theta_{y_i}+m)<cos(\theta_j)$.
Therefore the final formulation of the ArcFace loss can be written as:
\begin{eqnarray}
  L^{\mathrm{ArcFace}} =-\frac{1}{N} \sum_{i=1}^N \mathrm{log}\frac{e^{s\ \mathrm{cos}(\theta_{y_i}+m)}}
  {e^{s\ \mathrm{cos}(\theta_{y_i}+m)}  +\sum_{j=1,j\neq y_i}^C e^{s\ \mathrm{cos}(\theta_j)}}.
\end{eqnarray}

The margin-based methods decrease the probability of the ground truth class. It constrain the distribution of features strictly, that may cause overfitting.
\subsection{Center loss}
The center loss's intuition is to minimize intra-class variations. To achieve that, it decreases the euclidean distance between the feature $x_i$ and the class center $c_{y_i}$ directly, as formulated :
\begin{eqnarray}
  \label{eq:center1}
  L^C=\frac{1}{2 N} \sum_{i=1}^N \left \| x_i - c_{y_i}\right \|^2_2,
\end{eqnarray}
where $c_{y_i}$ denotes the $y_i$-th class center of the feature $x_i$. $c$ will converge to the center of the features of every class as the entire training set is taken into account. Besides, it needs the softmax loss to keep different classes separated. The final formulation is balanced the softmax loss (modified) and the center loss by a scale parameter $\lambda$: 
\begin{eqnarray}
  \label{eq:center2}
  \begin{split}
    L^{\mathrm{center}} &= \lambda L^C + L^{\mathrm{soft}} \\
    & = \frac{\lambda}{2 N} \sum_{i=1}^N \left \| x_i - c_{y_i}\right \|^2_2  -\frac{1}{N} \sum_{i=1}^N \mathrm{log}\frac{e^{W^T_{y_i} x_i}}{\sum_{j=1}^Ce^{W^T_{j} x_i}}.
  \end{split}
\end{eqnarray}

\section{METHOD}
\begin{figure}[!tb]
  \centerline{
    \includegraphics[width= .43\textwidth]{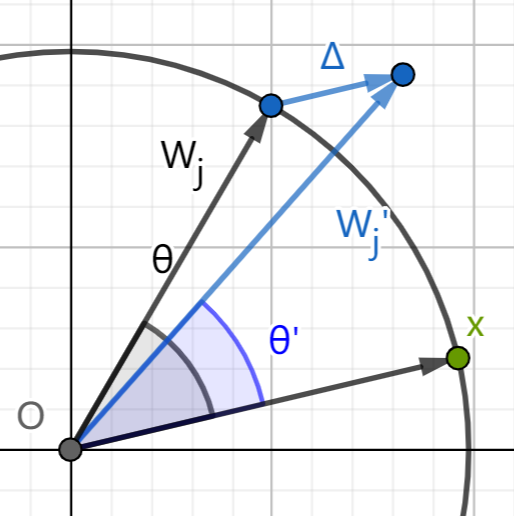}
  }
  \caption{A toy example of the process of updating $\theta$ during training, where $x'$ denotes the normalized feature, $W_j$ is normalized too and $\Delta$ is the gradient of last layer. $W_j'$ will become closer to $x$ if $j$ represents the ground truth label, $j==y$.}\label{fig:mothology}
\end{figure}

SphereFace \cite{liu2017sphereface} provides a new view on the weights of the last fully connected layer, representing the centers of each class in angular space.
Enlightened by this, we draw {a theory}---{minimizing the angle can achieve discriminative features and the fast way to decrease the angle $\theta$ is maximizing the gradient of $\theta$.}

In this paper, we propose a novel loss function named angular loss, which adds arccos after the last fully connected layer, to give a constant gradient.
\begin{eqnarray}
    L^{\mathrm{Arc}} =-\frac{1}{N} \sum_{i=1}^N \mathrm{log}\frac{e^{s\ \theta_{y_i}}}
  {\sum_{j=1}^C e^{s\ \theta_j}}.
  \label{eq:arc}
\end{eqnarray}

For convenience, we give a shorthand for each method, soft for the softmax loss, center for the center loss, ArcFace for the ArcFace loss, and Arc for the angular loss.

\begin{figure}[!th]
  \centerline{
    \includegraphics[width= .5\textwidth]{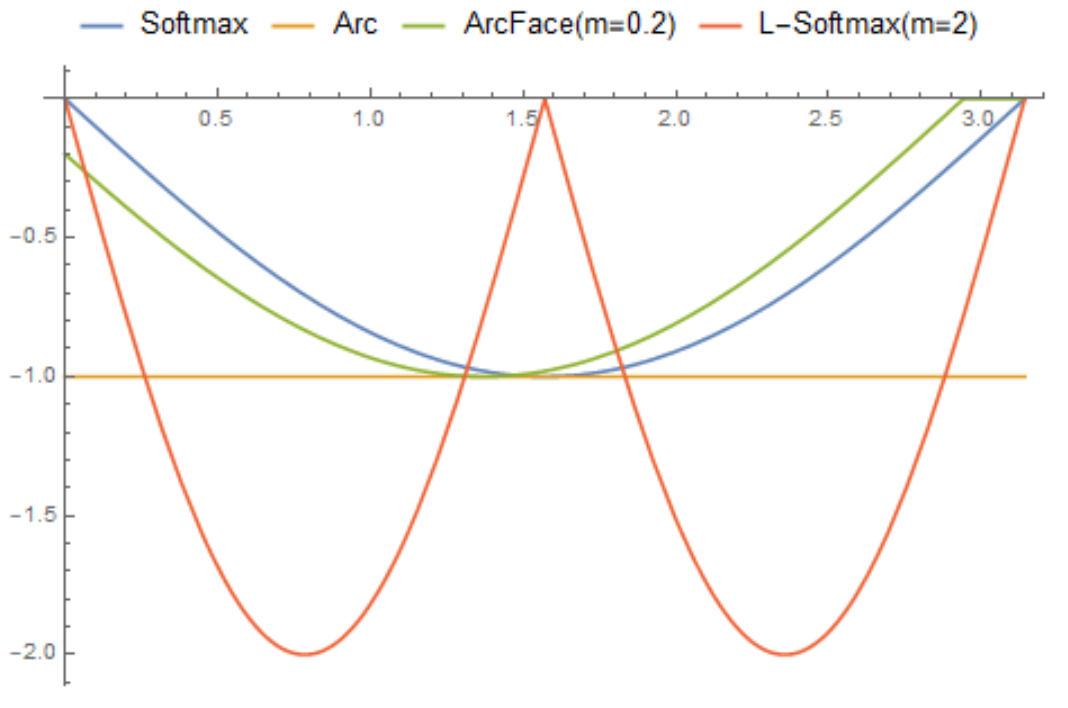}
  }
  \caption{The partial gradient $\frac{\partial Z}{\partial \theta }$ of four loss functions, including the softmax, ArcFace , L-Softmax \cite{liu2016large} and angular (our method) loss. $s=1$ for all these losses.}
  \label{fig:gradient}
\end{figure}
\subsection{Gradient analysis}
We investigate the change of the intra-class angle while training. To simplify this problem, we assume that the features $x$ remain unchanged on our analysis because the features are not updated directly by backpropagation. Figure \ref{fig:mothology} illustrates how $\theta$ is updated during training. 
The angular gradient indicates the change of the intra-class angle. Hence, our intuitive objective is to design a loss function which has a large angular gradient.

In softmax, the partial angular gradient is
\begin{eqnarray}
  \begin{split}
    \frac{\partial Z^{\mathrm{soft}}}{\partial \theta_j }&=\frac{\partial \left \| W^T_j \right \| \left \| x^i \right \| \mathrm{cos} (\theta_j)}{\partial \theta_j }\\
    &=-\left \| W^T_j \right \| \left \| x^i \right \| \mathrm{sin}(\theta_j).
  \end{split}
\end{eqnarray}

In ArcFace, the rescale parameter $s$ has no effect on gradient, and $ Z^{\mathrm{ArcFace}} = \left\{\begin{matrix}
    s\ \cos(\theta_j+m) & j=y_i     \\
    s\ \cos(\theta_j)   & j\neq y_i
  \end{matrix}\right.$. The derivation of $Z^{\mathrm{ArcFace}}$ is

\begin{eqnarray}
  \label{eq:g_arcface}
  \begin{split}
    \frac{\partial Z^{\mathrm{ArcFace}}}{\partial \theta_j } = \left\{\begin{matrix}
      -s\ \sin(\theta_j+m) & j=y_i     \\
      -s\ \sin(\theta_j)   & j\neq y_i
    \end{matrix}\right.
  \end{split}
\end{eqnarray}



The angular loss has a constant gradient, enhancing the intra-class compactness consistently.
\begin{eqnarray}
  \label{eq:g_arc}
  \frac{\partial Z^{\mathrm{Arc}}}{\partial \theta_j } =  -s.
\end{eqnarray}

Figure \ref{fig:gradient} compares the gradient of four loss functions.  In softmax, the gradient goes down to zero when $\theta$ is closing to zero. That means it is hard to optimize the intra-class compactness when $\theta$ is small. That ArcFace solves this problem by offset avoids the range of zero gradients. The sharp curve in L-Softmax leads to a smaller range of small angular gradient. Nevertheless, L-Softmax still has the zero angular gradient that makes it hard to train.

\begin{figure}[t] \centering    
\subfigure[The curve of $\frac{\partial L}{\partial \theta_{y_i}}$ w.r.t. $s$. The curve of $\frac{\partial L}{\partial \theta_{j}}$ is similar.] {
 \label{fig:gradient1}     
\includegraphics[width= .43\textwidth]{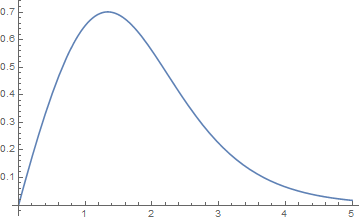}
}     
\subfigure[The curve of $\log(\frac{P(y_i)}{P(j)})$ w.r.t. $s$.] { 
\label{fig:ratio}     
\includegraphics[width= .43\textwidth]{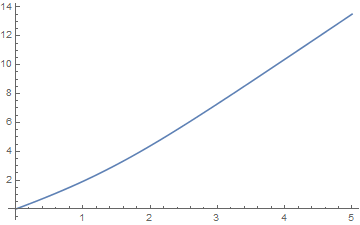}   
}    
\caption{The trade-off between a high angular gradient and high confidence.}     
\label{fig:trade-off}     
\end{figure}

\subsection{Adjustable constant}
Our method defines an extra constant $s$, which squeezes/stretches the $P(j)$ curvature. The angle between the feature and the mismatched class is close to $\frac{\pi}{2}$, similar to the situation in \cite{zhang2019adacos}.
Therefore, we assume that $\theta_{i,y_i}=0,\theta_{j,y_i}=\frac{\pi}{2}(j \neq y_i)$. $P(j)$ becomes
$$P(j)=\left\{\begin{matrix}
              \frac{e^{s\ \pi}}{e^{s\ \pi}+(C-1) e^{s\ \frac{\pi}{2}}} & j=y_i     \\
              \frac{e^{s\ \frac\pi 2}}{e^{s\ \pi}+(C-1) e^{s\ \frac{\pi}{2}}}   & j\neq y_i
    \end{matrix}\right.$$
    
First, we should ensure that $P(y_i)$ is far bigger than $P(j)$, that is, $\frac{P(y_i)}{P(j)}$ is very large. Empirically, $s$ should be greater than 3.Then, let's consider the angular gradient. According to the chain rule:
$$\frac{\partial L}{\partial \theta }= \frac{\partial L}{\partial Z } \frac{\partial Z}{\partial \theta },$$
combining EQ\ref{eq:g_arc}, the angular gradient of angular loss is
\begin{eqnarray}
\frac{\partial L^{\mathrm{Arc}}}{\partial \theta_j } =  \left\{\begin{matrix}
          -s\ (\frac{e^{s\ \pi}}{e^{s\ \pi}+(C-1) e^{s\ \frac{\pi}{2}}}-1) & j=y_i     \\
          -s\ \frac{e^{s\ \frac\pi 2}}{e^{s\ \pi}+(C-1) e^{s\ \frac{\pi}{2}}}   & j\neq y_i
\end{matrix}\right.
\end{eqnarray}

\begin{figure}
    \centering
    \includegraphics[width=.46\textwidth]{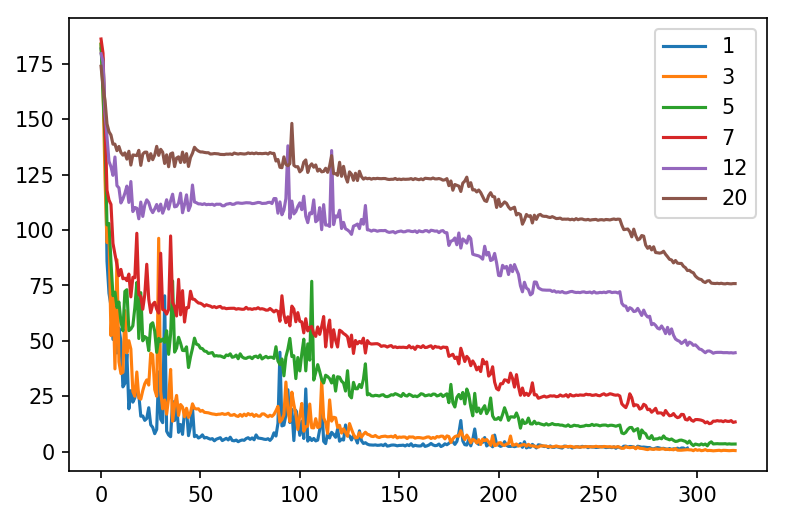}
    \caption{WC-Intra  angle vs. iteration. }
    \label{fig:angleVSs}
\end{figure}

Mathematically, the intra-class angle will keep decreasing when the angular gradient is bigger than zero. Therefore, the larger $s$ produces the smaller intra-class angle. We conduct the experiment on Fashion-MNIST comparing $s=1,3,5,7,12,20$ and detect the WC-Intra (Defined in EQ\ref{eq:Intra}) for every 200 iterations. The results in Figure \ref{fig:angleVSs} reveals that the constant $s$ adjusts the intra-class compactness. And $s$ increases the confidence of this class, as shown in Figure \ref{fig:trade-off}.

\section{EXPERIMENTS}
\subsection{General experimental settings}

\label{ds:description}
\textbf{Fashion-MNIST} \cite{xiao2017/online}: The Fashion-MNIST is a dataset of Zalando's article images designed for a replacement of the original MNIST\cite{lecun1998mnist}. Both are simple and light datasets, however, the most simple classification can reach 90\% test accuracy on MNIST. Many researchers have replaced MNIST to Fashion-MNIST in order to verify their ideas. \\
\textbf{CIFAR10/CIFAR10+} \cite{krizhevsky2009cifar}: The CIFAR10 dataset, containing ten classes, is widely used for image classification tasks. It is separated into a training set with 50000 samples and a test set with 10000 samples. CIFAR10+ denotes the CIFAR10 with data augment. For the data augmentation, we follow the transformations in \cite{lee2015deeply-supervised}: a $32\times32$ random cropping with 4-pixel padding on each side, a random flipping with the probability of 0.5, and a z-score normalization. \\
\textbf{CIFAR100/CIFAR100+} \cite{krizhevsky2009cifar}: We also testify our method on CIFAR100 dataset, which has the same image size but is more complex. Due to the similarity of CIFAR10 and CIFAR100, we remain our experiment setting almost unchanged. CIFAR100+ denotes CIFAR100 dataset with data augmentation.\\
\textbf{Architecture}: Deep residual networks \cite{he2016deep} have been widely used in image classification tasks \cite{hershey2017cnn,durand2017wildcat}, improving the performance of the deep convolutional neural networks. Though many other modern architectures, such as Inception-ResNet, WideResNet, ResNext \cite{khan2019survey}, are proposed. Our purpose is not achieving the best result of cifar10 but varifies the efficiency of our method. So we use the original ResNet and some modern training techniques introduced by Leslie Smith \cite{smith2018disciplined}. He also provides a practical approach to select hyperparameters (such as learning rate, weight decay, batch\_size). We follow this one cycle policy to change the learning rates. We summarize our experiments settings on Table \ref{table:setting}.
\begin{table}
  \centering
  \caption{The training settings of our experiments. ``CNU'' denotes the Convolution Units, and ``lr'' denotes the learning rate. Though the learning rate of these loss functions are different, all of them are well trained, reaching high accuracy (above 99.9\%) on training set. In all experiments, weight decay is 0.0005 and optimizer is Adam \cite{kingma2014adam}.}
  \label{table:setting}
  \begin{tabular}{|c|c|c|c|}
    \hline
    Dataset     & Fashion-MNIST & CIFAR10  & CIFAR100 \\ \hline
    CNU         & ResNet20       & ResNet18 & ResNet34 \\ \hline
    epochs      & 60            & 160      & 160      \\ \hline
    batch\_size & 256           & 128      & 128      \\ \hline
    lr(Soft)    & 0.1           & 0.01     & 0.01     \\ \hline
    lr(Arc)     & 0.1           & 0.1      & 0.1      \\ \hline
    lr(ArcFace) & 0.1           & 0.1      & 0.1      \\ \hline
    lr(Center)  & 0.01          & 0.01     & 0.01     \\ \hline
  \end{tabular}
\end{table}

\subsection{Angle descent verification}
In this section, we demonstrate the effectiveness of our method to descent the angle from different aspects. First, we use a toy example to plot the actual feature distribution; Second, we monitor the angles while training; Last, we give the confusion matrix to grasp the similarity of different features on real datasets (CIFAR10 and CIFAR100).
\subsubsection{Intuitive interpretation}

\begin{figure}[!b]
  \centering
  \subfigure[Softmax]{\label{fig:fashion_soft}
    \includegraphics[width=.22\textwidth]{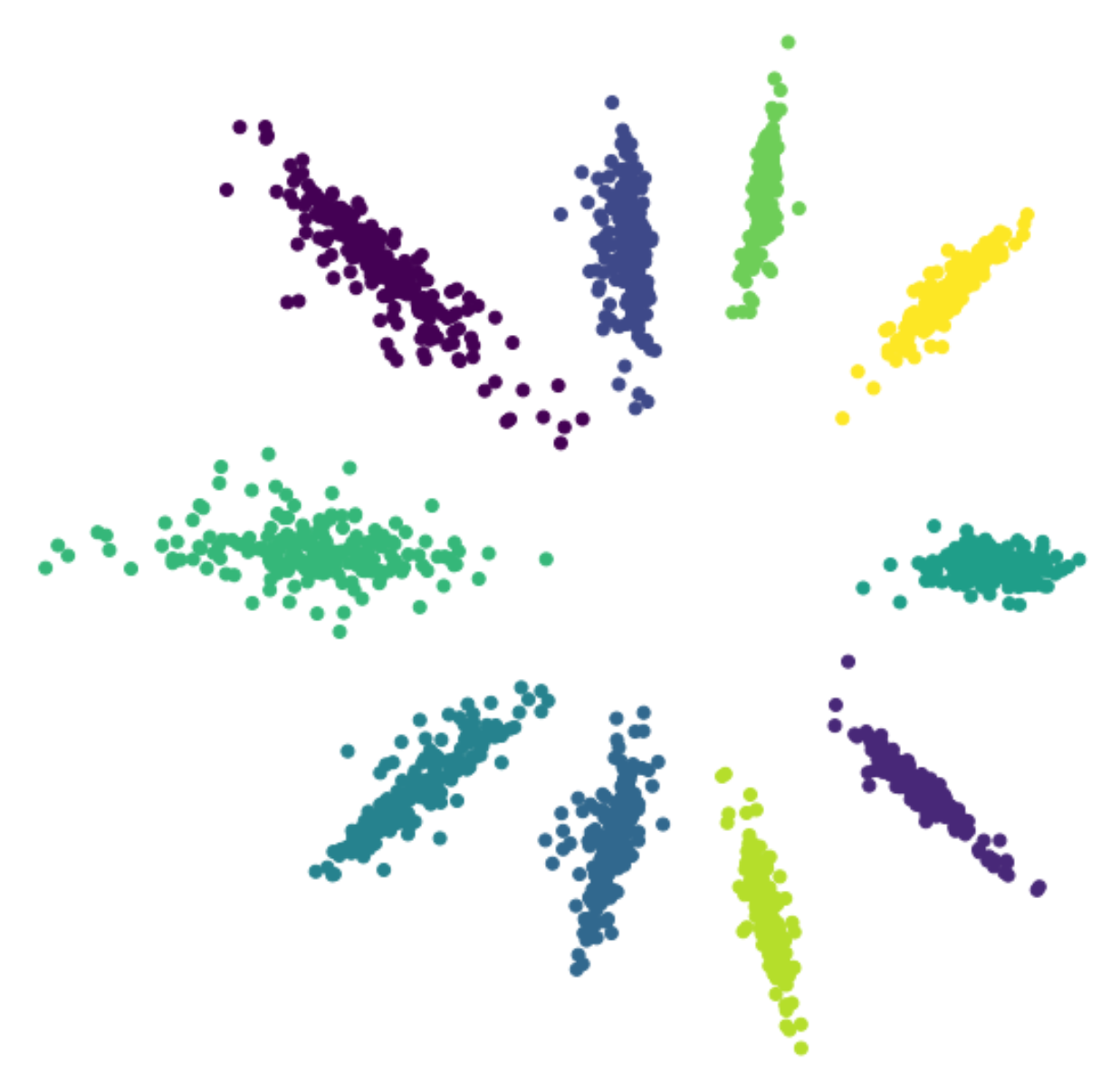}}
  \quad
  \subfigure[Center]{\label{fig:fashion_center}
    \includegraphics[width=.22\textwidth]{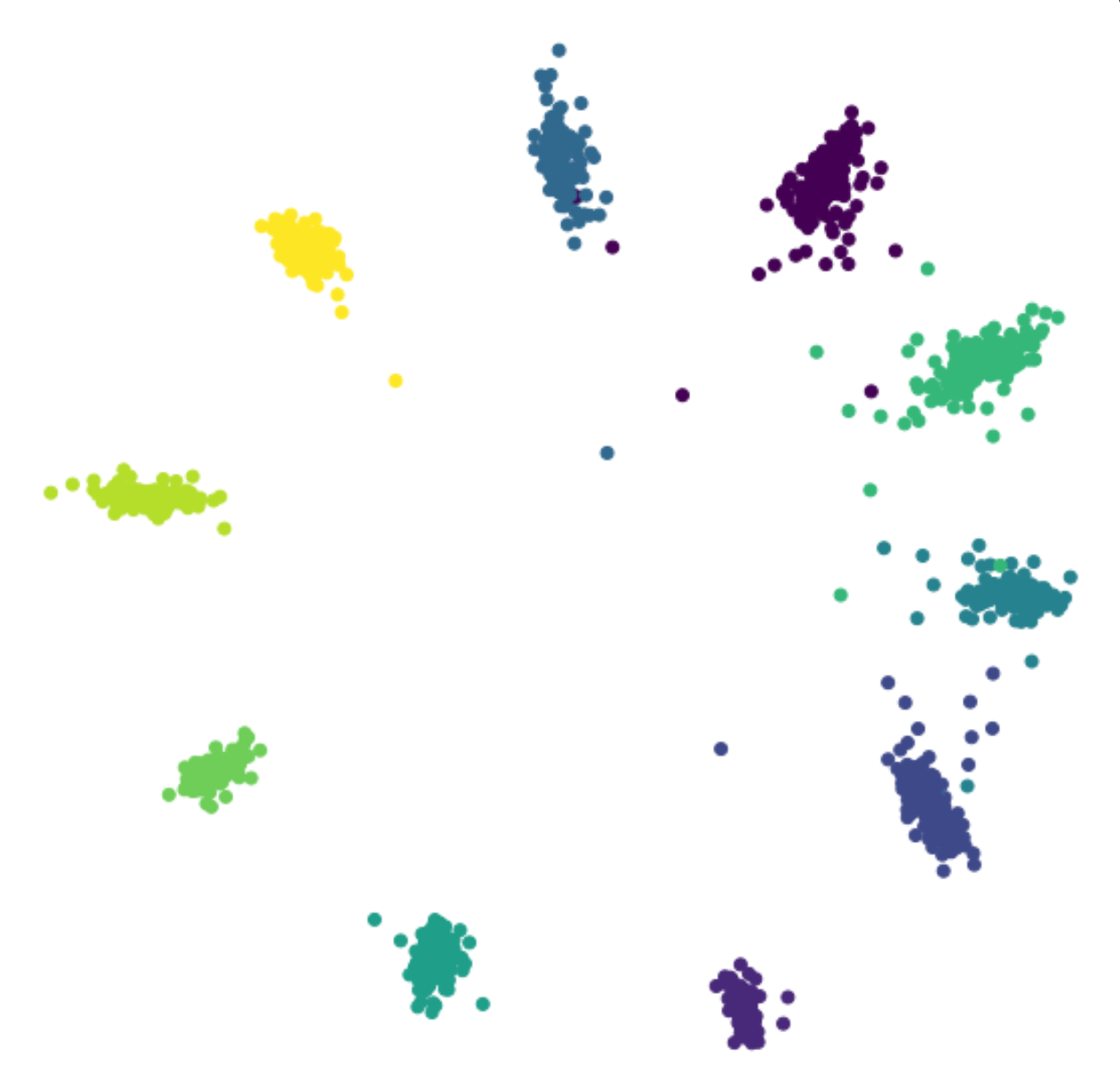}}\qquad

  \subfigure[ArcFace]{\label{fig:fashion_arcface}
    \includegraphics[width=.22\textwidth]{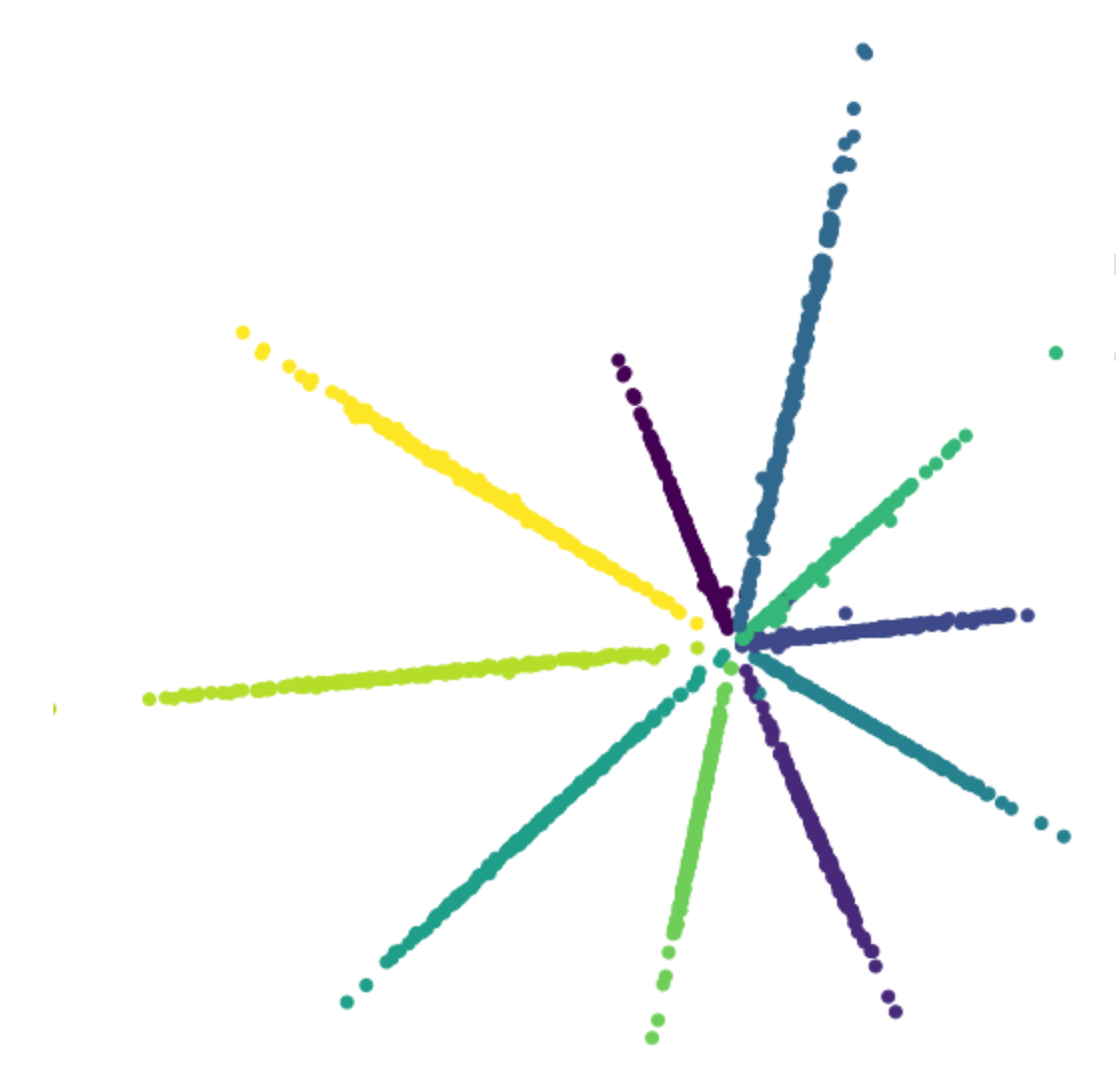}}
  \quad
  \subfigure[Arc]{\label{fig:fashion_arc}
    \includegraphics[width=.22\textwidth]{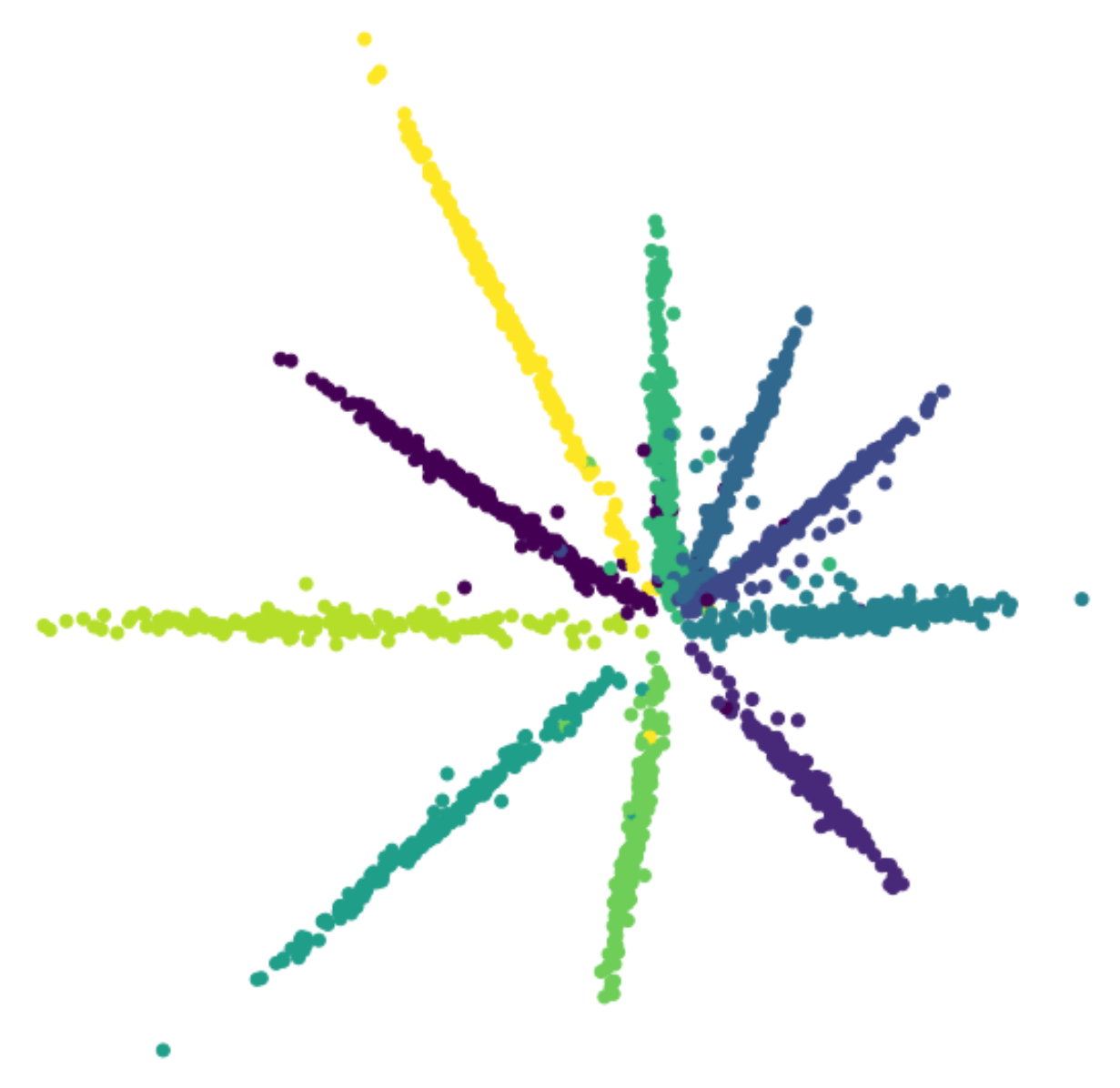}}\\
  \caption{Embedded features visualization on Fashion-MNIST dataset. Specially, we set the dimension of the features as 2. Though the features in Arc and ArcFace, nearing the origin of coordinates, seemed are close, they are far in angular space.}
  \label{fig:2d}
\end{figure}

We conduct this experiment on the Fashion-MNIST dataset. For visualizing the embedded features, we set the dimension of features as two and follow the training settings listed in Table \ref{table:setting}. Figure \ref{fig:2d} plots the distribution of features comparing the angular loss with the other loss functions, showing that the softmax loss generates roughly separable features, and other methods can produce discriminative features. ArcFace and Arc generate explicitly discriminative features; however, Arc is more tolerate on the outlier.

\subsubsection{Angle histogram}
\label{sec:angle}
Our method can directly optimize the angle between embedded feature and its target vector. To proof that, we define the following metrics \cite{deng2019arcface}:

\begin{flalign}
  \text{WC-Intra} &=  \frac{1}{C} \sum_{j=1}^C \langle \mathrm{Center}_j,W_j \rangle \\
  \text{W-Inter}  &=  \frac{1}{C^2-C} \sum_{j=1}^C \sum_{i=1,i\neq j}^C  \langle W_i,W_j \rangle \\
  \label{eq:Intra}
  \text{C-Inter}  &=  \frac{1}{C^2-C} \sum_{j=1}^C \sum_{i=1,i\neq j}^C \langle \mathrm{Center}_j,W_j \rangle
\end{flalign}
where $C$ is the number of class, $\mathrm{Center}_j = \bar{x}$ is the mean of the features belonging to the same class $j$, and $\langle a,b\rangle$ denotes the angle between a and b. ``W-Inter'' refers to the mean of angles between different target vectors $W^T_j$. ``C-Inter'' refers to the mean of angles between different classes' feature center. ``WC-Intra'' refers to the mean of the angles between target vectors $W^T_j$ and feature center of the class $j$. Table \ref{table:as1} and \ref{table:as2} give the details of angle statistics on CIFAR10 and CIFAR100+ dataset. To be attention, the intra-class angle of ArcFace is extraordinarily little, and that we argue that it probably causes overfitting.

\begin{table}
  \begin{center}
  {\caption{The angle statistics under different losses on CIFAR10+. ArcFace and Arc enlarge both inter-class separability and intra-class compactness.}\label{table:as1}}

  \begin{tabular}{|c|c|c|c|}
    \hline
    Method  & W-Inter         & C-Inter         & WC-Intra        \\ \hline
    Soft    & 1.2679          & 1.6366          & 0.9089          \\ \hline
    Arc     & \textbf{1.6821} & \textbf{1.6822} & 0.0039          \\ \hline
    ArcFace & \textbf{1.6821} & 1.6821          & \textbf{0.0021} \\ \hline
    Center  & 1.4749          & 1.6430          & 0.5604          \\ \hline
  \end{tabular}

  \end{center}
\end{table}

\begin{table}
  \centering
  \caption{The angle statistics (rad) under different methods on CIFAR100. ArcFace and Arc enlarge both inter-class separability and intra-class compactness.}
  \label{table:as2}
  \begin{tabular}{|c|c|c|c|}
    \hline

    Method  & W-Inter & C-Inter & WC-Intra \\ \hline
    Soft    & 1.5680  & 1.5648  & 0.4056   \\ \hline
    Arc     & 1.5404  & 1.5395  & 0.0301   \\ \hline
    ArcFace & \textbf{1.6256}  & \textbf{1.6244}  & \textbf{0.0103}   \\ \hline
    Center  & 1.5712  & 1.5601  & 0.3467   \\ \hline
  \end{tabular}
\end{table}

\begin{figure}[h]
  \center{
    \subfigure[Softmax]{\includegraphics[width= .4\textwidth]{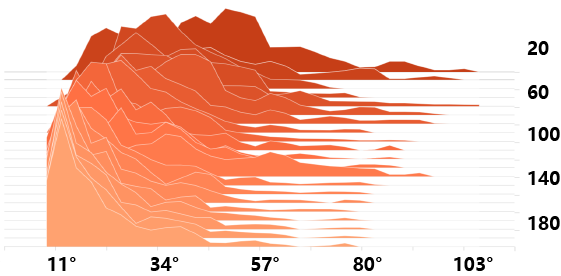}}\qquad
    \subfigure[Angluar Loss]{\includegraphics[width= .4\textwidth]{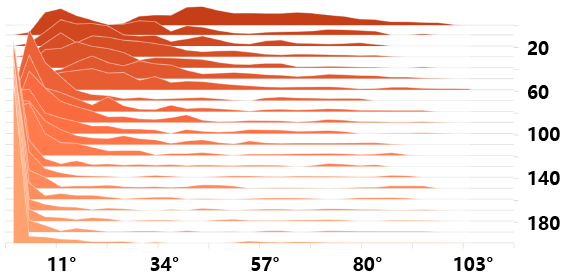}}
  }
  \caption{Visualization of the angle histogram in tensorboard \cite{tensorflow2015-whitepaper}. Each slice in the figure displays the angle histogram of intra-class. The slices depicts the change of angle during the training process. The top slice referring to the angle histogram at epoch 0 is darker and the lower slices referring to higher epoch are lighter. } \label{fig:distribution}
\end{figure}

We give the angle histograms of different classes to capture the dynamic change of the intra-class angles in Figure \ref{fig:distribution}. At the end of every epoch, we randomly pick 200 samples belonging to class 0  and measure the angle $\theta_0$ for each sample. In comparison with the softmax method, our approach will reduce the target angle faster and be at a lower level.

\begin{figure*}[tb]
  \centering
  \subfigure[Soft on CIFAR10]{
    \includegraphics[width=.2\textwidth]{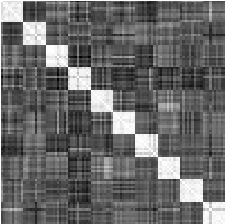}}
  \quad
  \subfigure[Center on CIFAR10]{
    \includegraphics[width=.2\textwidth]{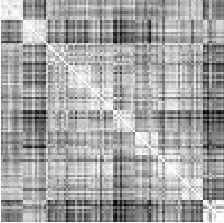}}
  \quad
  \subfigure[ArcFace on CIFAR10]{
    {\includegraphics[width=.2\textwidth]{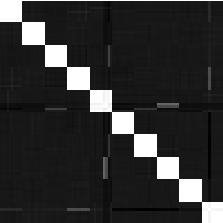}}}
  \quad
  \subfigure[Arc on CIFAR10]{
    \includegraphics[width=.2\textwidth]{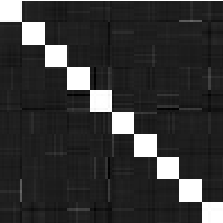}}\qquad

  \subfigure[Soft on CIFAR100]{
    \includegraphics[width=.2\textwidth]{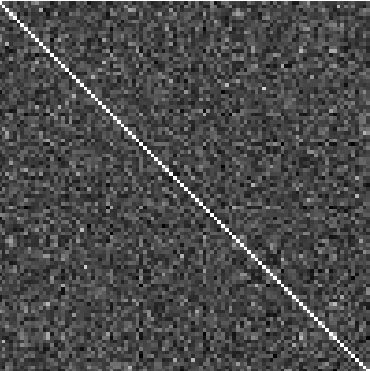}}
  \quad
  \subfigure[Center on CIFAR100]{
    \includegraphics[width=.2\textwidth]{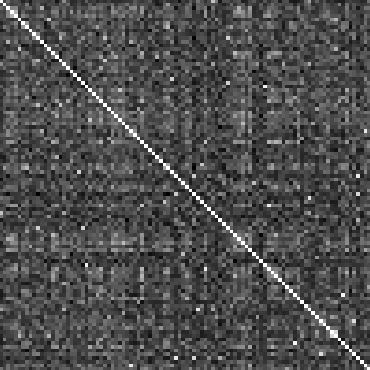}}
  \quad
  \subfigure[ArcFace on CIFAR100]{
    \includegraphics[width=.2\textwidth]{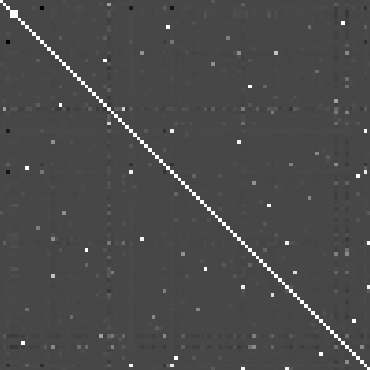}}
  \quad
  \subfigure[Arc on CIFAR100]{
    \includegraphics[width=.2\textwidth]{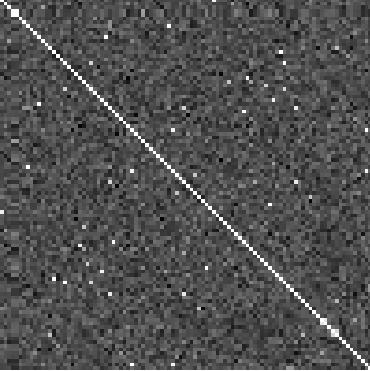}}

  \caption{The confusion matrix on CIFAR10 and CIFAR100 dataset.}
  \label{fig:CM}
\end{figure*}

\subsubsection{Confusion matrix visualization}
Visualization is difficult for high-dimensional features. Hence the comparison of the confusion matrix is given in  Figure \ref{fig:CM} to show the cosine similarity of features. In particular, on CIFAR10 dataset, we randomly select 10 features for each class, consisting of a total of 100 features; on CIFAR100 dataset, we randomly select one feature for each class. The learned features will then be applied to L2 standardization and calculated by
\begin{eqnarray}
    CM_{ij}= x_i^T \ x_j\ .
\end{eqnarray}
One can see that ArcFace and Arc greatly enhance intra-class compactness and enlarge the inter-class separability.

\subsection{Performance}
We have proved our method can enhance the intra-class compactness. All these experiments show that our method can significantly decrease the intra-class angle as our theory infers. In this part, we compare four loss functions: the softmax loss, the center loss, the ArcFace loss, and the angular loss (ours). 

The experiment on Fashion-MNIST, shown in Table \ref{table:fashion}, reveals our method and center loss are best. Table \ref{table:cifar10} and \ref{table:cifar100} reveal Arc is the best and ArcFace is better. That means discriminative information improves the accuracy, however ArcFace faces the problem of overfitting.

\subsection{Computational efficiency}
In this part, we use different epochs to train our model while keeping the other settings the same in Table \ref{table:setting}. Results in Table \ref{table:epochsVS.error} reveal our method accelerates training.
\begin{table}
  \centering
  \caption{The error rate(\%) vs. epochs. Every row indicates one experiment. Our method consistently outperforms others {within} the same training epochs.}
  \label{table:epochsVS.error}
  \begin{tabular}{ccccc}
    \hline
    Epochs & Soft   & Arc    & ArcFace & Center \\
    \hline
    10     & 15.72 & \textbf{12.09} & 14.08  & 17.67 \\
    20     & 12.55 &\textbf{ 8.65} & 9.23  & 10.73 \\
    80     & 9.43 & \textbf{6.07 }& 6.22  & 6.64 \\
    200    & 6.21 & \textbf{5.68} & 6.22  & 6.60 \\
    \hline
  \end{tabular}
\end{table}

\begin{table}
  \centering
  \caption{The error rate (\%) on Fashion-MNIST dataset.}
  \label{table:fashion}
  \begin{tabular}{|c|c|}
    \hline
    Method  & error\_rate   \\ \hline
    \hline
    Genetic DCNN \cite{MaAutonomousDCNN}        &   5.4  \\ \hline
    CNN  \cite{bhatnagar2017classification}     &   7.46 \\ \hline
    \hline
    Soft    & 6.30          \\ \hline
    \textbf{Arc}    & \textbf{6.04} \\ \hline
    ArcFace & 6.27          \\ \hline
    Center  & \textbf{6.04} \\ \hline
  \end{tabular}
\end{table}
\begin{table}
  \centering
  \caption{Recognition error rate (mean$\pm$std\%) on CIFAR10 without data augmentation and CIFAR10+ with data augmentation. Every result is evaluated five times.}
  \label{table:cifar10}
  \begin{tabular}{|c|c|c|c|}
    \hline
    Method                                  & Params(M) & CIFAR10             & CIFAR10+           \\ \hline
     \hline
    EM-Softmax \cite{wang2018soft-margin}   & 15.2 & -                 & 6.69        \\ \hline
    Maxout \cite{goodfellow2013maxout}      & -     & 9.38                & -               \\ \hline
All-CNN \cite{springenberg2014striving}     & 1.3      & 9.08                & 7.25               \\ \hline
     \hline
    Softmax                                 & 11.22    & 13.16$\pm$0.31          & 5.73$\pm$0.22          \\ \hline
    Center                                  & 11.22    & 12.22$\pm$0.45          & 5.40$\pm$0.21          \\ \hline
    \textbf{Arc}                                     & 11.22    & \textbf{11.77$\pm$0.31} & \textbf{5.23$\pm$0.19} \\ \hline
    ArcFace                                 & 11.22    & 12.11$\pm$0.45          & 5.34$\pm$0.12          \\ \hline
  \end{tabular}
\end{table}

\begin{table}
  \centering
  \caption{Recognition error rate (mean$\pm$std\%) on CIFAR100+ dataset with data augmentation. Every result is evaluated five times.}
  \label{table:cifar100}
    \begin{tabular}{|c|c|c|c|}
    \hline
    Method  & Params(M) & Top1                & Top5               \\ \hline
     \hline
    EM-Softmax \cite{wang2018soft-margin}   & 31.1 & 27.26                 & -        \\ \hline
    Maxout \cite{goodfellow2013maxout}      & -     & 38.57                & -               \\ \hline
All-CNN \cite{springenberg2014striving}     & 1.4      & 33.71                & -               \\ \hline
     \hline
    Softmax & 21.54    & 25.64$\pm$0.12          & 10.58$\pm$0.26         \\ \hline
    Center  & 21.54    & 25.33$\pm$0.96         & \textbf{8.19$\pm$0.58} \\ \hline
    \textbf{Arc}     & 21.54    & \textbf{24.29$\pm$0.08} & 10.07$\pm$0.23         \\ \hline
    ArcFace & 21.54    & 25.56$\pm$0.16          & \ 9.74$\pm$0.31         \\ \hline
    \end{tabular}
\end{table}

\section{CONCLUSION}
In this paper, we proposed an angular loss function, which has a constant $s$ to adjust the compactness of learned features. Our work also provides a potential direction to encourage intra-class compactness by the angular gradient analysis. Comparing to the hard constraint on margin-based methods, our method avoids overfitting by a soft way. The experiments also have demonstrated that our method outperforms the state-of-the-art. Moreover, our method can accelerate training.

\ack This work was supported by National Natural Science Foundation of China under Grant No. 61872419, No. 61573166, No. 61572230, No. 61873324, No. 81671785, No. 61672262. Shandong Provincial Natural Science Foundation No. ZR2019MF040, No. ZR2018LF005. Shandong Provincial Key R\&D Program under Grant No. 2019GGX101041, No. 2018GGX101048, No. 2016ZDJS01A12, No. 2016GGX101001, No. 2017CXZC1206. Taishan Scholar Project of Shandong Province, China.

\end{document}